\documentclass[letterpaper, 10 pt, conference]{ieeeconf}  
\usepackage{graphicx}

\IEEEoverridecommandlockouts                              

\usepackage{multirow}
\usepackage{xcolor}
\usepackage{comment}
\usepackage{url}
\usepackage{subcaption}
\usepackage{amsmath}
\usepackage{amssymb}
\usepackage{cite}
\usepackage{caption}
\usepackage{balance}
\title{\LARGE \bf
TIMID: Time-Dependent Mistake Detection in Videos of Robot Executions
} 
\makeatletter
\let\@oldmaketitle\@maketitle
\renewcommand{\@maketitle}{
  \@oldmaketitle
  \begin{center}
    \vspace{2em}
    \includegraphics[width=\linewidth]{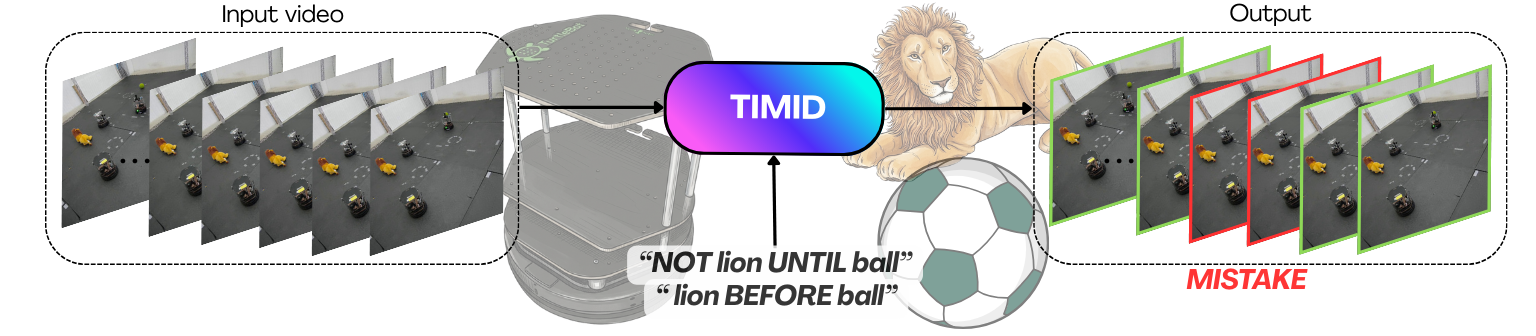}
    \vspace{-2em}
    \captionof{figure}{\textbf{TIMID:} proposed architecture to identify time-dependent mistakes in videos of Robot Executions. Our model takes a video and two prompts as input and outputs a mistake prediction at frame-level. The Video Anomaly Detection inspired architecture allows for weakly-supervised training using only video-level annotations.}
    \label{fig:teaser}
  \end{center}
  \vspace{-1em}
}
\makeatother

\author{Nerea Gallego$^{1*\dagger}$, Fernando Salanova$^{1*}$, Claudio Mannarano$^{1,2}$, Cristian Mahulea$^{1}$ and Eduardo Montijano$^{1}$
\thanks{*Equal contribution.}
\thanks{$^{\dagger}$Corresponding author.}
\thanks{This work was partially supported by grants AIA2025-163563-C31, PID2024-159284NB-I00, funded by MCIN/AEI/10.13039/501100011033 and ERDF; the Office of Naval Research Global grant N62909-24-1-2081; DGA project T45\_23R; and a 2024 DGA scholarship.}
\thanks{$^{1}$ Department of Systems Engineering and Computer Science, I3A, University of Zaragoza, Zaragoza, Spain
        {\tt\small ngallego@unizar.es}}%
\thanks{$^{2}$ University of Torino, Turin, Italy}%
}

\begin{document}

\maketitle
\setcounter{figure}{1}
\thispagestyle{empty}
\pagestyle{empty}

\begin{abstract}
As robotic systems execute increasingly difficult task sequences, so does the number of ways in which they can fail.
Video Anomaly Detection (VAD) frameworks typically focus on singular, low-level kinematic or action failures, 
struggling to identify more complex temporal or spatial task violations, because they do not necessarily manifest as low-level execution errors.
To address this problem, the main contribution of this paper is a new VAD-inspired architecture, TIMID, which is able to detect robot time-dependent mistakes when executing high-level tasks.
Our architecture receives as inputs a video and prompts of the task and the potential mistake, and returns a frame-level prediction in the video of whether the mistake is present or not.
By adopting a VAD formulation, the model can be trained with weak supervision, requiring only a single label per video.
Additionally, to alleviate the problem of data scarcity of incorrect executions, we introduce a multi-robot simulation dataset with controlled temporal errors and real executions for zero-shot sim-to-real evaluation.
Our experiments demonstrate that out-of-the-box VLMs lack the explicit temporal reasoning required for this task, whereas our framework successfully detects different types of temporal errors.

Project: \url{https://ropertunizar.github.io/TIMID/}
\end{abstract}

\section{INTRODUCTION}

Recent advances in large-scale imitation learning~\cite{chi2025diffusion} and foundation models~\cite{pmlr-v305-black25a} have expanded the perceptual and behavioral capabilities of robots, enabling them to execute complex task sequences with minimal human supervision. However, as task complexity scales, correctness can no longer be assessed solely at the level of individual actions. Instead, overall success depends on whether the full execution remains consistent with the high-level task description. 

Current autonomous frameworks still lack explicit task-level and temporal awareness. While modern policies can often recover from minor physical perturbations, they typically do not \emph{recognize} when an execution has deviated from the intended procedure. Importantly, many task failures are \emph{not} kinematic outliers: a robot can execute a visually correct action (e.g., grasping or approaching a target) yet violate the high-level goal because it happens at an incorrect stage or under an unmet precondition. We refer to these as \emph{time-dependent mistakes}~\cite{taxonomy}, violations of temporal constraints over task predicates, even if each atomic action is correct.

Within the computer vision community, procedural analysis methods~\cite{flaborea2024prego,huang2025modeling} often rely on rigid graph-based representations of the task, where states and transitions must be explicitly defined and manually annotated.
In contrast, Video Anomaly Detection (VAD) methods~\cite{pel4vad} require only weak supervision, typically labeling entire videos as either correct or anomalous. However, traditional VAD approaches focus primarily on explicit anomalies, such as traffic accidents or security breaches, which manifest as clear spatial or kinematic deviations.
In this paper, we demonstrate that VAD methods, when adapted to robotic demonstrations, can move beyond only detecting visual outliers, identifying different time-dependent mistakes using only coarse demonstration-level labels during training.

To validate this claim, the paper makes two contributions. First, we introduce TIMID a VAD-inspired architecture for detecting time-dependent mistakes from weakly labeled video demonstrations. The model takes as input a task and mistake descriptions and a video of robot execution, and produces frame-level mistake predictions using only video-level supervision during training (Fig.~\ref{fig:teaser}).
Second, we present a formally generated multi-robot simulation dataset for studying time-dependent execution errors. The dataset supports training under controlled mistake scenarios and includes real robot executions for evaluating sim-to-real performance.

\section{LITERATURE REVIEW}

\subsection{Error Detection in Robot Executions}
Ensuring the reliable execution of robotic tasks has driven extensive research across various sub-domains of robotics. Historically, error detection has been highly compartmentalized based on the specific type of failure. In mobile robotics, significant effort has been dedicated to navigation errors, such as collision detection~\cite{park2021collision, ji2022proactive} and path deviations~\cite{sorbelli2020uavs}. Similarly, in Human-Robot Interaction (HRI), error detection typically focuses on safety violations, social navigation failures, or unintended physical contacts \cite{trung2017head, haddadin2017robot, heo2019collision}. 

In robotic manipulation, monitoring has traditionally relied on non-visual modalities. Some works demonstrated the detection of anomalous robot motion in collaborative manufacturing by tracking kinematics and IoT sensor data \cite{zhong2023detecting}. Other approaches utilize multimodal sensory feedback (e.g., force-torque sensors or tactile feedback) to identify anomalies like slipping or failed grasps \cite{stachowsky2016slip, saxena2006robotic, yoo2021multimodal}. Other works detect high-level deviations at task level utilizing global trajectory logs, resulting more similar to data mining rather to anomaly detection \cite{salanova2025highlevel}. While recent efforts incorporated vision to predict basic manipulation anomalies based on optical flow, these methods remain strictly limited to short-horizon, localized physical failures \cite{thoduka2021using}. 

Despite the variety of execution monitoring literature, there is a lack of methods capable of identifying high-level, temporal semantic errors in complex, multi-step tasks. Existing frameworks evaluate \textit{how} an action is performed, but fundamentally fail to evaluate \textit{when} or \textit{why} it is performed within the broader context of the task. 

\subsection{Video Anomaly and Mistake Detection}
In the computer vision community, Video Anomaly Detection (VAD) has been extensively studied, though its application to robotics remains sparse. Standard VAD frameworks~\cite{sultani2018real, ren2021deep, nayak2021comprehensive, nguyen2019anomaly} are predominantly designed for surveillance, excelling at identifying explicit, visually obvious anomalies, such as explosions, accidents, or fights. These methods usually categorize events based on visual chaos rather than task execution~\cite{abdalla2405video}.
To leverage the semantic power of Vision-Language Models (VLMs) for anomaly detection, architectures like PEL4VAD~\cite{pel4vad} integrate text prompts with video features for general anomaly contexts.
Other works have explored error detection in procedural tasks~\cite{huang2025modeling}, but they heavily rely on rigid, hard-coded task graphs limited to egocentric human views~\cite{flaborea2024prego, lee2024error}.

In this work we adopt a procedural mistake taxonomy~\cite{taxonomy}, but use a VAD methodology to detect them, simplifying the required supervision to train our model from a full task graph to a weak label at the video layer.

\subsection{Datasets for Robotic Anomalies}
The advancement of temporal anomaly detection in robotics is severely hindered by a data generation bottleneck. 
Existing single-scene VAD datasets~\cite{ramachandra2020survey} are saturated with pedestrian anomalies~\cite{ramachandra2020street, liu2018ano_pred}. 
In robot learning, large-scale datasets like BridgeData V2~\cite{walke2023bridgedata} provide extensive nominal demonstrations of single-arm manipulation. While valuable for benchmarking localized executional mistakes, these datasets do not inherently contain structured anomalies. 
Existing datasets capturing complex tasks and their corresponding time-dependent compliance failures are mostly centered on ensembling and cooking tasks made by humans~\cite{grauman2024ego}.

Regarding robotic environments, multi-robot high-level demonstrations are entirely absent from the current literature, motivating the novel dataset proposed in this work.

\section{METHODOLOGY}

\subsection{Problem Formulation}

Let a video be $F = \{f_1,f_2,\cdots,f_T\}$, where $f_t$ with $t=1, \ldots, T$ represents the visual frame at time step $t$, and a textual description, $\mathcal{P},$ of the task the robots are supposed to perform and the potential mistake, $\mathcal{M}$. 
The objective is to learn a scoring function $f(F, \mathcal{P}, \mathcal{M}) \rightarrow \{\hat{y}_t\}_{t=1}^T$, that indicates whether the mistake is present at time $t$ or not.

\subsection{Mistake Modeling}
In order to define the different types of mistakes, we adopt an existing formal taxonomy~\cite{taxonomy} that classifies them 

in two disjoint sets, $\mathcal{M} = \mathcal{M}_{exec} \cup \mathcal{M}_{proc}$.

\emph{Executional mistakes}, $\mathcal{M}_{exec}$, localize physical deviations where an expected action is performed incorrectly, e.g., failed grasp, slippage, incorrect contact.

\emph{Time dependent, or procedural, mistakes}, $\mathcal{M}_{proc}$, are protocol-level deviations where actions may be individually correct but violate temporal or logical task constraints, e.g., performing steps in the wrong order, skipping prerequisites or violating mutual exclusion.
To characterize this type of mistakes, we model the task description using Linear Temporal Logic (LTL) formulas~\cite{saha2014automated}. 

A task specification is expressed as a conjunction of LTL formulas,
\begin{equation}
    \varphi = \bigwedge_k \varphi_k.
\end{equation}
A time dependent mistake occurs when the actions of the agents in the scene violate $\varphi$ and can also be specified by another LTL formula that is in conflict with the correct one.

This formal description of the mission and the mistakes has the additional advantage of being close to natural language, enabling its integration with current VLM pipelines.
Particularly, our method simply needs textual prompts of the task and mistake to work, as shown in the example of Fig.~\ref{fig:teaser}.

\subsection{VAD architecture for Temporal Mistake Detection}

\begin{figure}[!htb]
     \includegraphics[width=\linewidth]{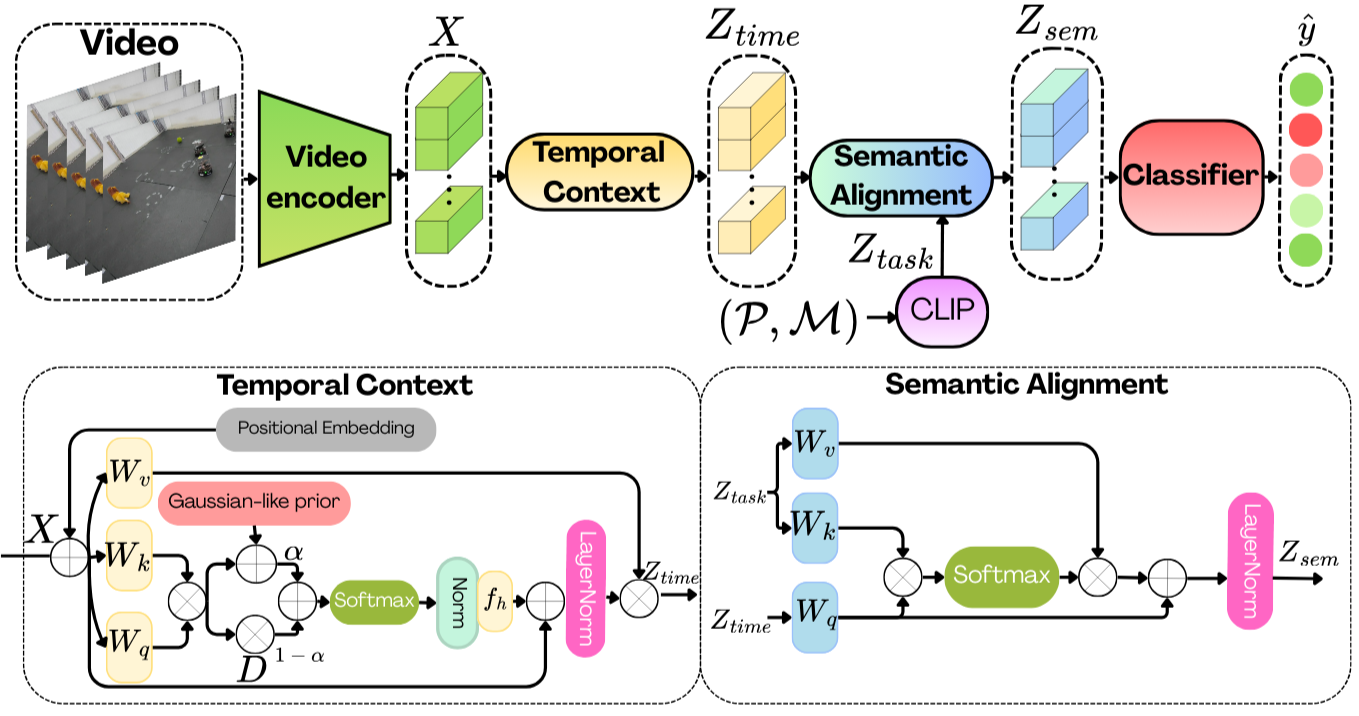}
     \caption{\textbf{Overview of the proposed time-dependent mistake detection pipeline.} The system processes video streams, tasks, and mistake descriptions to identify semantic and temporal deviations from high-level task objectives.}
     \label{fig:pipeline}
 \end{figure}

In order to detect the temporal mistakes in the videos, we propose TIMID, a VAD-inspired architecture that takes the video, $F$, the task, $\mathcal{P},$ and mistake, $\mathcal{M}$, descriptions as inputs and outputs if the execution has mistakes or not and where it fails.
The overview of the proposed architecture is shown in Fig.~\ref{fig:pipeline}. It is composed of a video encoder and two attention blocks, one used to learn the temporal context of the different embeddings and the other to align them with the semantic concepts of the task and the mistake.

\paragraph{Video Encoder} Instead of processing the entire video at once, we split it into non-overlapping fragments using a sliding window. 
Then, each segment is passed through a pre-trained video backbone to convert the raw information, $f_t$, into high-level feature vectors $X$. 

\paragraph{Temporal Context} 

Inspired in a previous VAD method ~\cite{pel4vad}, our model includes a temporal context module to simultaneously learn local and global temporal contexts. 
Unlike the module in~\cite{pel4vad}, ours includes a standard sinusoidal Positional Encoding directly to the input features to establish the absolute temporal order of the sequence. 
Then, a learnable, Gaussian-like prior, $G = \exp(-|\gamma(i - j)^2 + \beta|)$, is used as a second dynamic position encoding to account for the instant in which each visual feature happens. 
Last, the module integrates this prior into the standard Query-Key-Value formulation. The similarity matrix is computed as $\mathcal{E} = \frac{QK^T}{\sqrt{d_k}}+G$, where $d_k$ is the dimensionality of the queries. To capture both bidirectional context and temporal dependencies, the module employs a dual-stream architecture: global and local stream. The global stream computes an unmasked context $C_{global}=Softmax(\mathcal{E})V$, while the local stream computes a causal context $C_{local}=Softmax(\mathcal{E} \odot D)$, where $D$ is a lower-triangular mask preventing attention to future frames.
Finally, these two context representations are fused using a learnable scalar parameter to balance the global and causal information:
\begin{equation}
    Z_{time} = \sigma(\alpha)C_{global}+(1-\sigma(\alpha))C_{local},
\end{equation}
where $\sigma$ denotes the sigmoid activation function.
Differently from~\cite{pel4vad}, we consider the entire video context during training but each feature only takes into account the visible information up to that point.

\paragraph{Semantic Alignment} 

To bridge the gap between raw visual data and task description, our architecture includes a semantic reasoning module designed to map ``bad executions'' into a shared latent space.

We use a pretrained CLIP text encoder~\cite{radford2021learning} to extract semantic features, $Z_{task},$ from the task and mistake prompts, $\mathcal{P}$ and $\mathcal{M}$. 

Our framework employs a cross-attention mechanism to align textual rules temporally within the video.
Being $Z_{time}$ the temporal features, we project them into Queries ($Q$), and the text features, $Z_{task},$ into Keys ($K$) and Values ($V$),
$$Q = Z_{time} W_Q, \quad K = Z_{task} W_K, \quad V = Z_{task} W_V,$$
where $W_Q$, $W_K$ and $W_V$ are learnable linear projection matrices.

The module learns to attend to specific spatial-temporal regions corresponding to task violations by computing the scaled dot-product attention,
$$\text{Context} = \text{Softmax}\left(\frac{QK^\top}{\sqrt{d_k}}\right)V.$$

To ensure stable optimization and retain the original visual information, we apply a residual connection followed by Layer Normalization,$$Z_{sem} = \text{LayerNorm}(\text{Context} + Q).$$

\paragraph{Classifier} 
Finally, a linear projection $W_O$ maps the aligned representations to the final output dimension, $\hat{y}_t = Z_{sem} W_O$, resulting in a frame-level scoring value about the presence of the mistake.

\paragraph{Training and testing procedures} 
Our framework operates under a strictly weakly supervised paradigm. During training, the model only has access to video-level labels indicating whether a mistake occurred anywhere in the sequence. However, at inference, the model is capable of generating fine-grained, frame-level anomaly predictions. To achieve this, we train the network using a joint loss function $\mathcal{L}=\mathcal{L}_{bce}+\mathcal{L}_{con}$.

To enable weakly supervised temporal prediction using only video-level labels, we formulate the classification as a Multiple Instance Learning (MIL) problem~\cite{lv2023unbiased}. Given a sequence of frame-level mistake logits $S$ for a video of length $T$, we dynamically pool these scores into a single video-level representation, $s_{pool}$, based on the ground truth. For normal videos, we extract the maximum frame score ($s_{pool} = \max(S)$) to strictly penalize any false alarms. For anomalous videos, we average the top-$k$ highest scores—where $k = \max(1, \lfloor T/32 \rfloor)$, to capture the specific temporal feature where the failure occurs.  Optimizing this pooled score via a Binary Cross-Entropy with Logits loss ($\mathcal{L}_{bce}$) forces the model to localize mistakes to specific time without requiring dense frame-level annotations.

To further separate the feature space of failure modes, we apply a contrastive loss ($\mathcal{L}_{con}$). We first isolate the valid frames (discarding padding) and apply mean-pooling across the temporal dimension to generate a single global video representation, $f_{global} = \frac{1}{T} \sum_{t=1}^T f_t$. A supervised contrastive loss is then applied to these global features to cluster videos with similar labels while pushing apart normal and anomalous representations.

At test time, the model directly outputs the frame-level mistake scores without requiring the pooling operations.

\section{DATASET}
\label{sec:dataset}

A recurrent problem to train VAD models is the lack of incorrect demonstrations compared to normal examples, which in the context of robotics and time-dependent mistakes, $\mathcal{M}_{proc}$, is even more aggravated.
To address this,

in the paper we also introduce a new dataset of robotic collaborative tasks aimed to 
cover the higher end in the taxonomy~\cite{taxonomy}. 

\subsection{Task descriptions}
Our dataset considers a team of robots operating within a physical arena with two objects of interest, a lion plush and a green ball. We define two atomic propositions $\varphi_1=\texttt{Lion}$ and $\varphi_2=\texttt{Ball}$, which are active whenever a robot is in the vicinity of the objects.
The dataset focuses on two task that require semantic and temporal behavior: 
\begin{itemize}
    \item Mutual exclusion: the robots cannot visit simultaneously the lion and the ball. The constraints is expressed as 
    $$\varphi_{\text{mutex}}=\mathbf{G} \ \neg \left( \texttt{Lion} \land \texttt{Ball}\right),$$ 
    where $\mathbf{G}$ denotes ``globally'' operator, requiring the condition to hold at all time steps.
    The prompts used for describing the task and anomaly are $\mathcal{P}=$``robot NOT IN lion AND green ball'' and $\mathcal{M}=$``robot IN lion AND green ball''.
    \item Sequential Ordering: robots need to visit the ball before visiting the lion,
    $$\varphi_{\text{order}}=\neg \texttt{Lion} \ \mathbf{U} \ \texttt{Ball},$$ where $\mathbf{U}$ denotes the ``UNTIL'' temporal operator. 
    In this case, the prompts used for describing the task and anomaly are $\mathcal{P}=$``robot NOT IN lion UNTIL in green ball'' and $\mathcal{M}=$``robot IN lion BEFORE green ball''.
\end{itemize}

\begin{figure}[!h]
    \centering
    \includegraphics[width=0.75\linewidth]{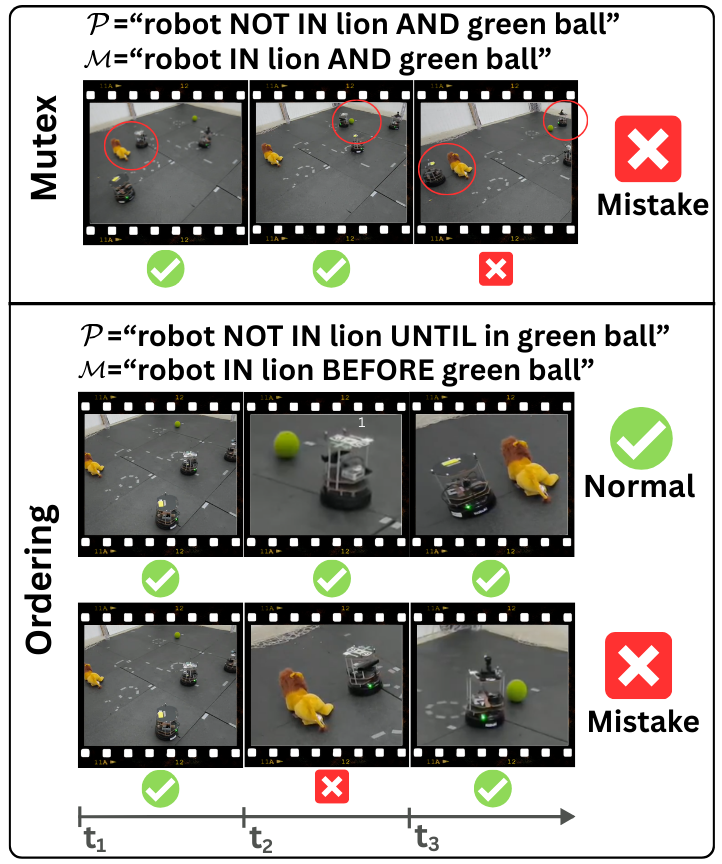}
    \caption{Description of the tasks contained in the dataset. The top-half shows a mutual exclusion task, focused on concurrency. The lower-half shows an ordering task, with emphasis on time. The dataset includes frame and video-level annotations.}
    \label{fig:datasetAnomaly}
\end{figure}

Figure~\ref{fig:datasetAnomaly} represents visually examples of the tasks contained in the dataset.

\subsection{Video generation}

\begin{figure*}[t] 
    \centering
    
    \begin{subfigure}[b]{0.28\linewidth}
        \centering
        \includegraphics[width=\linewidth, height=4cm]{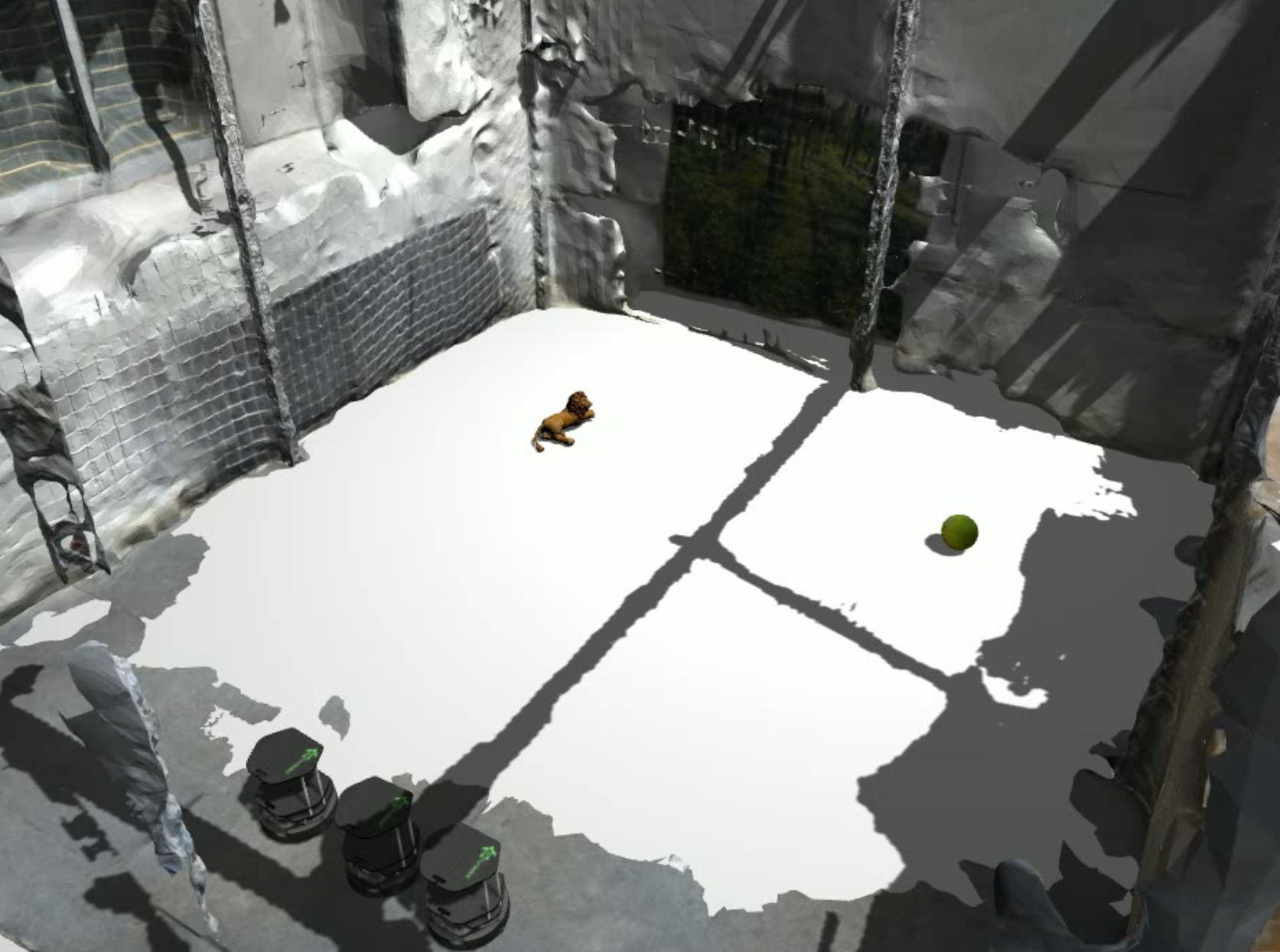} 
        \caption{Side View}
        \label{fig:pov1}
    \end{subfigure}
    \hfill 
    \begin{subfigure}[b]{0.28\linewidth}
        \centering
        \includegraphics[width=\linewidth, height=4cm]{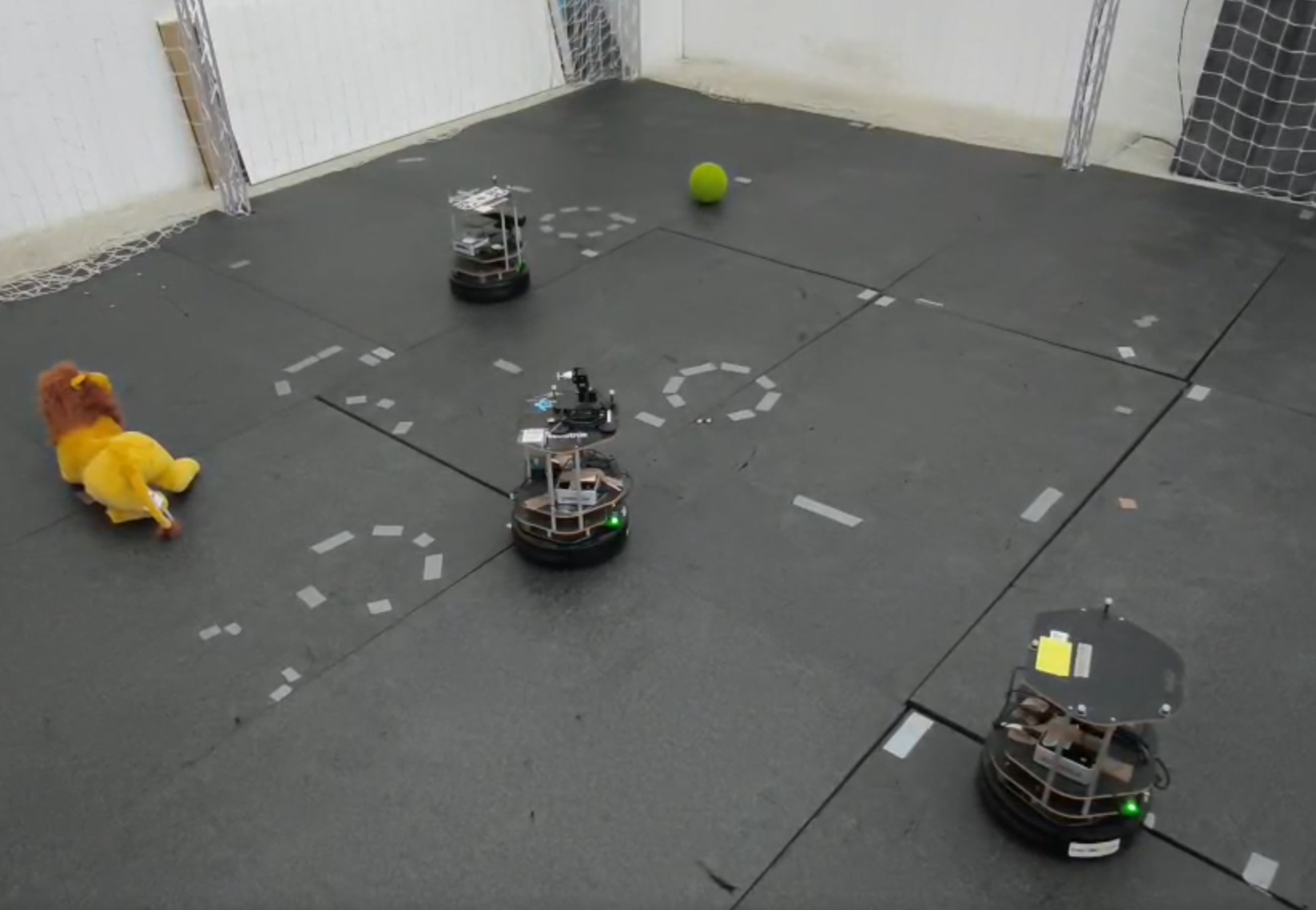} 
        \caption{Real Video}
        \label{fig:pov2}
    \end{subfigure}
    \hfill
    \begin{subfigure}[b]{0.28\linewidth}
        \centering
        \includegraphics[width=\linewidth, height=4cm]{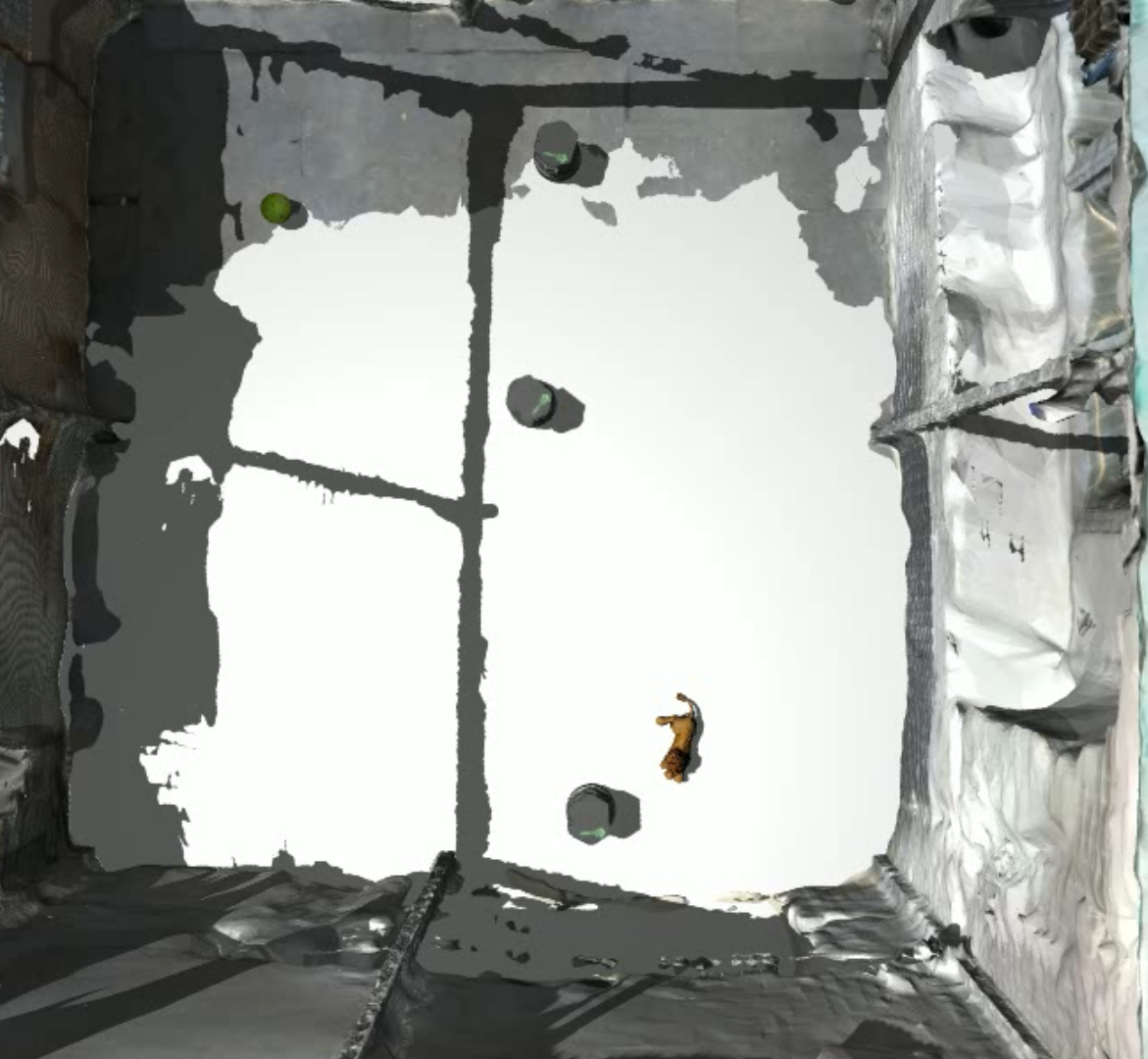} 
        \caption{Cenital View}
        \label{fig:pov3}
    \end{subfigure}
    
    \caption{Different examples of frames in the multi-robot dataset, captured across multiple points of view and localizations.}
    \label{fig:dataset}
\end{figure*}

Generating real videos of multiple robots performing different variations of the tasks, as well as executions containing mistakes, is very costly and difficult.
Therefore, in our dataset we have opted for the Gazebo simulation environment to enable fast and automatic generation of any arbitrary number of videos with different configurations.
Nevertheless, the whole simulation has been designed in a way that we have included in the dataset real executions to test the sim-to-real capabilities of the different models.
Figure~\ref{fig:dataset} shows a display of examples contained in the dataset.

\paragraph{Environment generation}
We have used a photo-realistic reconstruction in Gazebo of our experimental arena (Fig.~\ref{fig:dataset} a), where we have also included models of three real Turtlebots and the two objects.
We have considered three different spatial distributions of the objects within the arena and the robots always start from a depot location, with some initial variations in their locations.

\paragraph{Atomic action plans}
In order to generate individual actions for the robots, we start from the LTL formulas, $\mathcal{P}$ and $\mathcal{M}$, and generate a Büchi automaton to represent the task as a graph~\cite{saha2014automated}.
Then, we leverage the use of the \emph{Renew} simulator~\cite{kummer2004extensible} to produce any number of different sequences of actions that are always compliant with the task~\cite{HUSTIU2026105287}.
With this approach, we can work with an arbitrary number of robots and assign them different roles in different episodes, e.g., the robot that visits the lion changes from one video to another.
Moreover, since the tasks do not require the use of three robots, the action plans include additional unrelated actions that act as decoys in the videos.
Finally, modeling the mistakes as another formula makes the generation of both good and bad executions transparent. 
In total, we have generated $25-30$ different action plans per configuration (Environment+Task).

\paragraph{Low level execution}
In order to bridge the atomic action plans with the low-level real motion of the robots trajectories we have used the navigation stack in ROS2 (Nav2),
which is also the low-level controller used by the real platforms.

\paragraph{Video recordings}
Each execution has been recorded from three different camera points of view: a cenital view and two opposing side views (Fig.~\ref{fig:dataset} a and c). 
This leads to a dataset containing over 1000 annotated simulated videos across the different tasks and object distributions.
Additionally, the dataset contains $8$ videos of the real robots (Fig.~\ref{fig:dataset} b) from a single point of view and object configuration, $2$ correct and $2$ with mistakes for each task.

\subsection{Annotations}

To directly support the dual-prediction training paradigm of our architecture, the dataset is annotated at two distinct granularity using binary labels (i.e., Mistake Present / No Mistake Present). 
First, each sequence contains a \textit{video-level} annotation indicating if the execution failed to carry out the task at any point.
Second, the videos are uniformly divided into discrete temporal segments, and annotations are provided for every 16 frames. 
This high-resolution temporal localizes at fine-grained detail when the multi-robot team deviated from the task.

\section{EXPERIMENTS \& RESULTS}

To evaluate our proposed framework and demonstrate its capacity to span the full robotic anomaly taxonomy ($\mathcal{M}_{exec}$ and $\mathcal{M}_{proc}$), we structured our experimental setup into defined benchmarks, baselines, and metrics. This evaluation is designed to isolate the impact of temporal reasoning, test the limits of modern semantic models, and assess real-time applicability.

\subsection{Benchmarks}
We evaluate the models across two distinct data environments. 

To benchmark localized, physical errors ($\mathcal{M}_{exec}$), we utilize the BridgeData V2 dataset (Bridge) \cite{walke2023bridgedata}. This environment features real videos of a single robotic arm performing atomic manipulation actions in a kitchen setting (e.g., handling knives, pots, and ingredients). With it we aim at testing the ability to detect short-horizon semantic anomalies, such as a robot incorrectly grabbing food directly instead of using the proper kitchen utensil.

In order to evaluate high-level, time-dependent protocol violations ($\mathcal{M}_{proc}$), we use the multi-robot dataset introduced in Section~\ref{sec:dataset}.

In Bridge we have downloaded the first $1000$ episodes and manually labeled them based on the simple task of grabbing food, normal execution, and not grabbing it as the mistake.
We have randomly selected 20\% of the episodes and include fine-grained annotations every 16 frames for test.
In the Multi-robot dataset we have used 80\% of the simulation videos for training and used the remaining 20\% as well as the 8 real sequences for test.
All the models are trained independently for each type of mistake.

\subsection{Baselines}
We compare our architecture (TIMID) against different anomaly detection algorithms:
(i) A traditional LSTM-based autoencoder~\cite{sangeethapriya2024time} (Auto-Encoder) trained via a semi-supervised approach to flag temporal anomalies based purely on reconstruction errors. It serves as baseline for spatial-outlier detection entirely devoid of semantic awareness; (ii) A Vision-Language Model featuring 7 billion parameters~\cite{hui2024qwen2} (Qwen 2.5) deployed in a zero-shot capacity to tests whether massive, pre-trained semantic knowledge is sufficient to detect temporal mistakes directly from video prompts without domain-specific training; (iii) the same VLM has been fine-tuned (Qwen 2.5 ft) using the training split of each benchmark; and (iv) an existing VAD model~\cite{pel4vad} (PEL4VAD) trained in the same conditions as ours.

\subsection{Metrics}

To evaluate the models, we measure standard detection metrics, Average Precision (AP), Average Recall (AR) and F1, computed at frame level. 
We also report inference time for the whole dataset, measured in minutes.

All the experiments were performed on a workstation equipped with an AMD Ryzen 9 9950X processor, an NVIDIA GeForce RTX 5090 graphics card, and 64 GB of RAM. 

\subsection{Results}
Table \ref{tab:main_results} shows the results of all the baselines across the benchmarks.

\begin{table}[ht]
\caption{Mistake detection results}
\label{tab:main_results}
\centering
\footnotesize
\renewcommand{\arraystretch}{1.1} 
\setlength{\tabcolsep}{0.5em}
    \begin{tabular}{|l|c|c|c|c|c|}
    \hline
    \textbf{DATA} & \textbf{Model} & \textbf{AP} & \textbf{AR} & \textbf{F1} & \textbf{Inf.Time (m)}\\ \hline
    
    \multirow{5}{*}{Bridge} 
    & Auto-Encoder~\cite{sangeethapriya2024time}  & 9.33 & 8.09 & 8.67 & 1.16 \\ \cline{2-6}
    & Qwen 2.5~\cite{hui2024qwen2}& 58.21 & 29.8 & 39.42 & 119.7 \\ \cline{2-6}
    & Qwen 2.5 (ft)& \textbf{59.50} & \textbf{44.87} & \textbf{51.16} & 100.3 \\ \cline{2-6}
     & PEL4VAD~\cite{pel4vad} & 10.18 & 19.29 & 13.33 & \textbf{0.02}\\ \cline{2-6}
    & TIMID (Ours)  & 49.72 & 33.77 & 40.22 & \textbf{0.02} \\ \cline{2-6}
    \hline
    \hline
    
    \multirow{5}{*}{Mutex} 
    & Auto-Encoder~\cite{sangeethapriya2024time}  & 37.55 & 10.90 & 16.89 & 0.63 \\ \cline{2-6}
    & Qwen 2.5~\cite{hui2024qwen2}& 35.60  & 22.71 & 27.37 & 423.6  \\ \cline{2-6}
    & Qwen 2.5 (ft) & 41.48 & 24.41 & 30.73 & 473.2   \\ \cline{2-6}
     & PEL4VAD~\cite{pel4vad} & 64.43 & 35.20 & 45.53 &  0.03 \\ \cline{2-6}
    & TIMID (Ours)  & \textbf{76.83} & \textbf{35.89} & \textbf{49.1} & \textbf{0.02} \\ \cline{2-6}
    \hline
    \hline
    
    \multirow{5}{*}{Ordering} 
    & Auto-Encoder~\cite{sangeethapriya2024time}  & 11.80 & 5.04 & 7.06 & 0.02 \\ \cline{2-6}
    & Qwen 2.5~\cite{hui2024qwen2}& 28.92 & 7.46 & 11.86 & 403.5 \\ \cline{2-6}
    & Qwen 2.5 (ft) & 17.44 & 11.81 & 14.08 & 409.5  \\ \cline{2-6}
    & PEL4VAD~\cite{pel4vad} & 24.15 & 15.55 & 18.92 & \textbf{0.01}  \\ \cline{2-6}
    & TIMID (Ours)  & \textbf{48.71} & \textbf{36.89} & \textbf{41.98} & 0.02 \\ \cline{2-6}
    \hline

    \end{tabular}

\end{table}

The high-level takeaways from this evaluation show that video anomaly detection models can be used for high level mistake recognition.

On the Bridge dataset, where further temporal or logical reasoning is not needed, Qwen 2.5 offers the best results. 
This demonstrates that massive parameter spaces, are highly capable of parsing short-horizon, localized physical errors. 
However, TIMID achieves highly competitive predictive accuracy in this domain.

The limitations of existing baselines become evident in the multi-robot benchmarks. When tasked with tracking strict spatial mutual exclusions (Mutex) or sequential dependencies (Ordering), the general-purpose VLMs fail to maintain the historical multi-agent context required to identify rule violations. 
In contrast, TIMID dominates across all accuracy metrics on these high-level tasks. 
Another fundamental limitation of using a VLM is the time, which represents a huge bottleneck in contrast to the fast inference of TIMID and the other baselines. 

\begin{figure*}[!ht]
    \centering
    \includegraphics[width=0.45\linewidth]{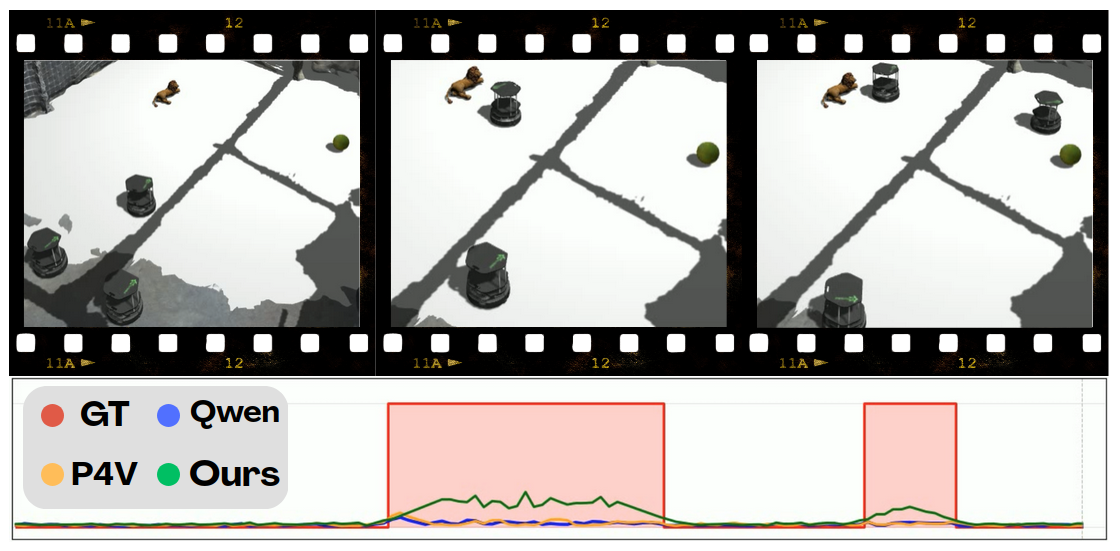}
    \includegraphics[width=0.45\linewidth]{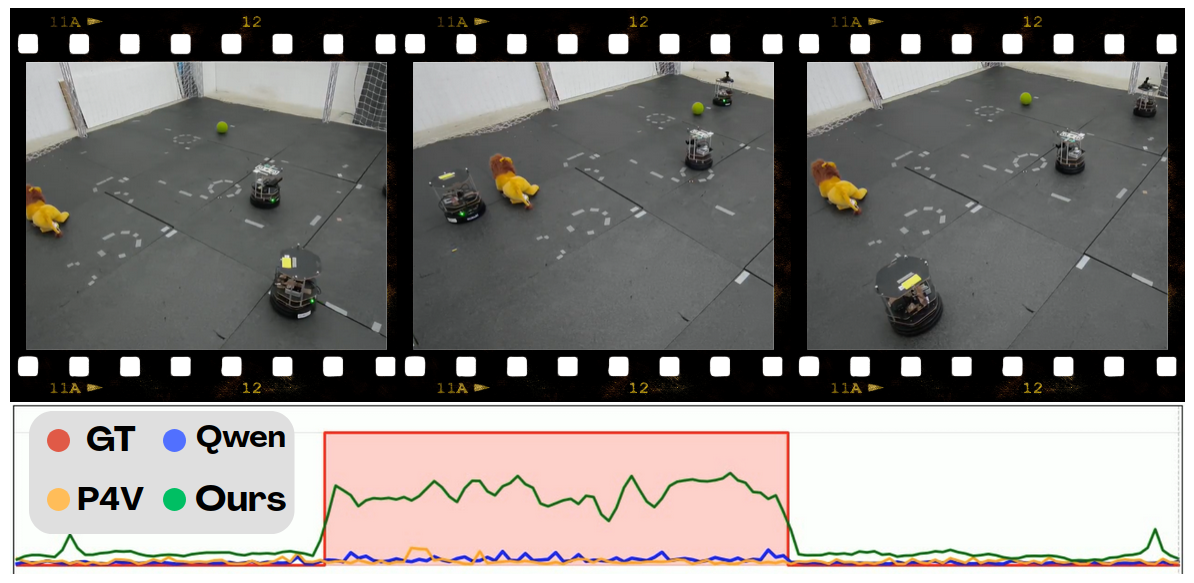}
    \includegraphics[width=0.45\linewidth]{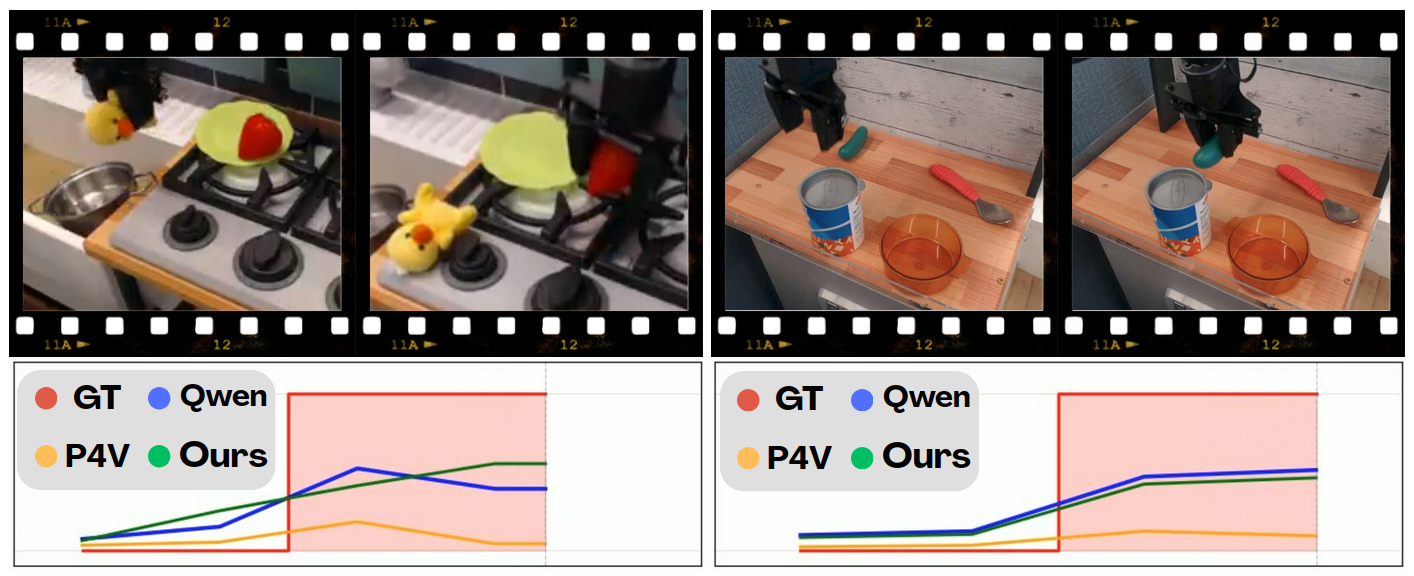}
    \includegraphics[width=0.45\linewidth]{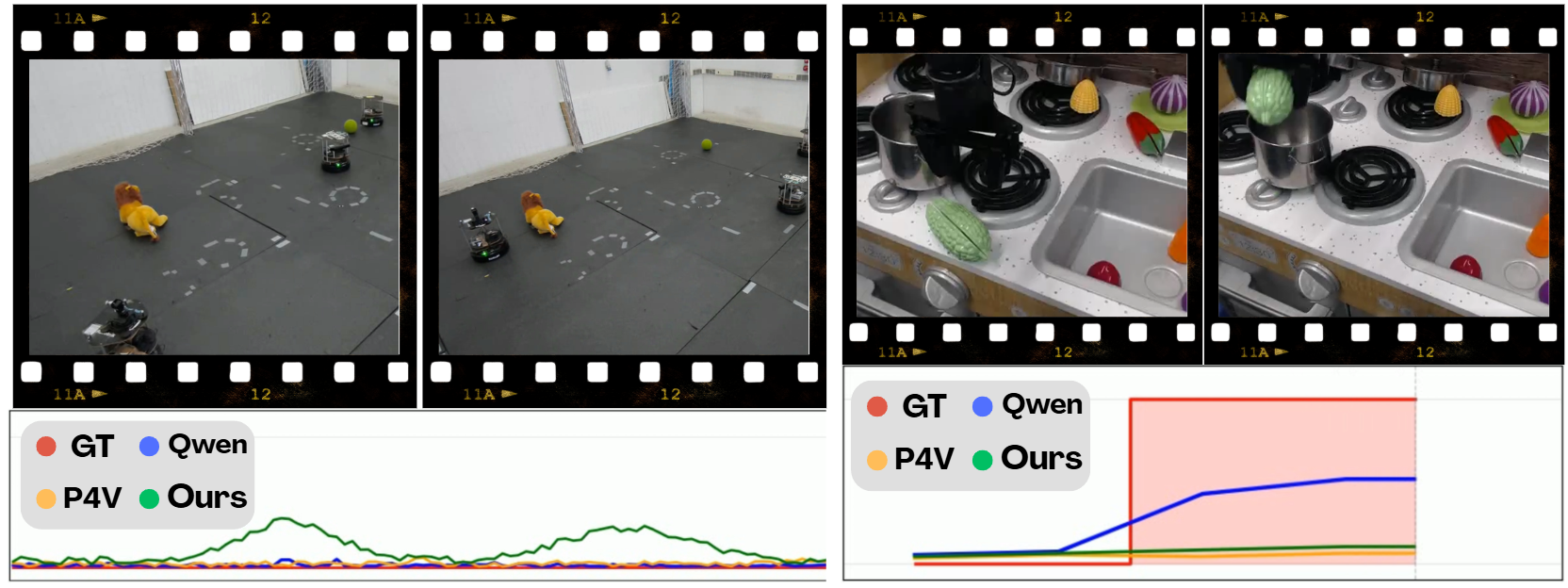}
    \caption{\textbf{Figure with examples of multiple predictions of the models across the different benchmarks.} On top of the predictions, frames of the execution exemplyfing the videos. Below the frames are the predictions of our model (in green) and the different baselines (Qwen in blue and PEL4VAD in yellow) are shown against the ground truth (in red). (The top left example shows a example of a synthetic execution of the ordering case, top right a real video of the proximity case, down left examples of the bridge dataset and down right some incorrect executions with examples of false positives and negatives).}
    \label{fig:Examples}
\end{figure*}

Figure~\ref{fig:Examples} includes different qualitative examples of the predictions made by the different models, including some failure cases of TIMID.

\subsection{Simulation to real}

Models trained on synthetic or simulated data often suffer severe performance degradation when exposed to real world data during deployment.

To evaluate the performance of our architecture against this domain shift, we designed a zero-shot sim-to-real experiment. 
In this phase, models trained exclusively on our synthetic datasets are tested with the real-world video sequences without any additional fine-tuning. 

As demonstrated in Table~\ref{tab:real}, crossing the reality gap drops the quality of the results for all the models.
Nevertheless, TIMID demonstrates more resilience to this domain shift than its competitors, specially in the precision of its predictions. 
This showcases that our approach does not merely memorize simulated visual layouts, but is able to learn the underlying semantics of the task and mistake.

\begin{table}[ht]
\caption{ZeroShot in real videos}
\label{tab:real}
\centering
\small
\renewcommand{\arraystretch}{1.1} 
\setlength{\tabcolsep}{0.5em}
    \begin{tabular}{|l|c|c|c|c|}
    \hline
    \textbf{Data} & \textbf{Model} & \textbf{AP (\%)} & \textbf{AR (\%)} & \textbf{F1} \\ \hline
    
    \multirow{5}{*}{Real Videos} 
    & Auto-Encoder & 16.15 & 9.14 & 11.67 \\ \cline{2-5}
    & PEL4VAD & 12.69 & \textbf{19.99} & 15.53 \\ \cline{2-5}
    & Qwen 2.5 & 14.47  & 12.05 & 13.15 \\ \cline{2-5}
    & Qwen 2.5 (ft) & 16.08 & 11.17 & 13.18  \\ \cline{2-5}
    & Timid (Ours)  & \textbf{45.94} & 18.88 & \textbf{26.76} \\ \cline{2-5}
    \hline
    
    \end{tabular}

\end{table}

\subsection{Ablation Studies}

Lastly, we conduct an ablation study to validate our architectural design choices. 

We isolate the core modules, \emph{Temporal} and \emph{Semantic Only}, of our pipeline to observe their individual impacts across the benchmarks, detailed in Table \ref{tab:ablation_results}.
While the two modules obtain competitive results independently, with \emph{Temporal Only} even beating the full model in some individual metrics, the joint use of them is what provides the best overall (F1 score) results in all datasets. 
Results also corroborate that each module its performing its intended function, as observed with the better results of \emph{Temporal Only} than \emph{Semantic Only} on the Ordering task and viceversa for Mutex.

\begin{table}[ht]
\caption{Ablation Study of Module impact}
\label{tab:ablation_results}
\centering
\small
\renewcommand{\arraystretch}{1.1} 
\setlength{\tabcolsep}{0.5em}
    \begin{tabular}{|l|l|c|c|c|}
    \hline
    \textbf{Dataset} & \textbf{Model} & \textbf{AP} & \textbf{AR} & \textbf{F1 Score} \\ \hline
    
    \multirow{3}{*}{Bridge} 
    & All (Ours) & \textbf{49.72} & 33.77 & \textbf{40.22} \\ \cline{2-5}
    & Semantic Only & 30.70 & 29.87 & 30.28 \\ \cline{2-5}
    & Temporal Only & 33.13 & \textbf{34.69} & 33.89 \\ \hline
    \hline

    \multirow{3}{*}{Mutex} 
    & All (Ours) & \textbf{76.83} & \textbf{35.89} & \textbf{49.1} \\ \cline{2-5}
    & Semantic Only & 68.31 & 24.12 & 35.65 \\ \cline{2-5}
    & Temporal Only & 52.59 & 24.74 & 33.65 \\ \hline
    \hline
    
    \multirow{3}{*}{Ordering} 
    & All (Ours) & 48.71 & \textbf{36.89} & \textbf{41.98} \\ \cline{2-5}
    & Semantic Only & 26.94 & 27.97 & 27.45 \\ \cline{2-5}
    & Temporal Only & \textbf{55.66} & 31.13 & 38.93 \\ \hline
    
    \end{tabular}
\end{table}

\section{LIMITATIONS}

While the proposed framework successfully bridges semantic reasoning and temporal tracking, it has some limitations that outline future work. 
The current architecture is trained to detect singular mistakes over a specific task, requiring to be re-trained each time they change. 
The extension of the model to classify multiple concurrent anomalies might help overcoming this.
Additionally, although we only need weak video-level supervision, our training strategy still relies on examples of 
anomalous executions, which are difficult to obtain outside a highly controlled environment, and might not be possible in certain scenarios.
The use of unsupervised techniques, like process-mining, might be a way to transition to purely unsupervised training on normal videos.

\section{CONCLUSIONS}

In this work, we studied the problem of detecting time-dependent mistakes in robotic task executions from videos. 
We proposed a VAD-inspired architecture that leverages task and mistake textual descriptions to produce frame-level predictions while being trained with only video-level supervision. 
The model combines a video encoder with attention modules that capture temporal context and align visual features with task and mistake semantics, detecting 
procedural errors in formally defined high-level tasks without explicitly encoding the task structure.
We also introduced a multi-robot simulation dataset with controlled temporal violations and real robot executions for sim-to-real evaluation. 
Experiments on this dataset and on real manipulation videos show that vision-language models struggle to detect procedural mistakes, supporting the potential of VAD-based approaches for this problem. 
Future work will focus on improving generalization, enabling multi-anomaly detection, and reducing supervision by training with only normal videos.

\balance

\bibliographystyle{IEEEtran}

\end{document}